\documentclass[conference]{IEEEtran}
\IEEEoverridecommandlockouts
\usepackage{cite}
\usepackage{amsmath,amssymb,amsfonts}
\usepackage{graphicx}
\usepackage{textcomp}
\usepackage{xcolor}
\usepackage{booktabs}
\usepackage{multirow}
\usepackage{url}
\def\BibTeX{{\rm B\kern-.05em{\sc i\kern-.025em b}\kern-.08em
    T\kern-.1667em\lower.7ex\hbox{E}\kern-.125emX}}

\begin{document}

\title{Semantically Rich, Metrically Affect-Blind:\\ A Geometric Blind Spot in Modern Text Embedders}

\author{\IEEEauthorblockN{Mahdi Naser Moghadasi}
\IEEEauthorblockA{\textit{BrightMind AI}\\
Seattle, WA, USA \\
mahdi@brightmind-ai.com}}

\maketitle

\begin{abstract}
We report a systematic dissociation in modern text embedders: sentiment is
almost perfectly \emph{decodable} from their embeddings, yet almost entirely
absent from their \emph{metric}. Across eleven frozen encoders spanning 2021 to
2026---including 2026-generation models (Microsoft Harrier-OSS-v1, Jina-v5-text)
and the 4B-parameter LLM embedder Qwen3-Embedding-4B---a linear probe recovers
sentence polarity at up to $91.3\%$ accuracy while cosine similarity separates
polarity by a margin of only $0.003$--$0.064$. Qwen3-Embedding-4B, the strongest
probe in the panel, reaches $91.3\%$ probe accuracy but only $64.3$ retrieval
MAP. Because retrieval, clustering, deduplication, and
retrieval-augmented generation all operate on the metric rather than on a trained
probe, these systems are effectively blind to information their encoder
demonstrably contains. A task-conditioning control suggests this reflects the
pretraining objective rather than a prompting artifact: conditioning Jina-v5 on its \emph{classification} task
rather than its \emph{retrieval} task recovers only about half of the MAP gain
that injection achieves on the same weights ($63.2 \rightarrow 73.8$, versus
$85.1$), and leaves the alignment gap essentially unchanged. We then show the dissociation is cheaply repairable. Sentiment-Injected
Sentence Embeddings (SISE) train a small (0.13M--0.69M-parameter)
supervised-contrastive head on a frozen encoder in seconds on a CPU, using the
labelled training set. The head improves every encoder released since 2022 on
full-data probe accuracy (up to $+5.8$ points), retrieval MAP (up to $+26.0$) and
polarity margin ($7$--$61\times$). This is a specialization, not a free gain: the
compact $128$-d head loses $38$--$41$ Spearman points on STS-B, while a fused
variant that concatenates it with the base embedding loses $2.4$--$12.0$. With
the head, a $0.6$B encoder at $128$ dimensions matches a $4$B encoder given the same
head ($91.8$ vs.\ $90.3$ MAP), so for sentiment retrieval encoder scale matters
far less than the metric. No corpus re-encoding is needed. We release a
CPU-reproducible pipeline and a reusable diagnostic, the Attribute Alignment Gap.
\end{abstract}


\section{Introduction}
Text embeddings are the substrate of large-scale text analytics. Retrieval,
clustering, deduplication, recommendation, and retrieval-augmented generation
(RAG) all reduce to nearest-neighbour search in an embedding space, which means
that in deployment what matters is not what a classifier could \emph{learn} to
read out of a vector, but what the \emph{metric} on that space actually
expresses. This distinction is usually invisible, because for topical and
semantic content the two coincide: encoders trained with contrastive objectives
place semantically related sentences close together, and probes and cosine
similarity agree.

This paper reports a case where they sharply disagree. We measure sentiment
polarity in eleven frozen encoders spanning five years of progress---from a 2021
MiniLM through E5, BGE, mxbai, and Snowflake Arctic-Embed, to 2025--2026
state-of-the-art models including Qwen3-Embedding (0.6B and 4B), Microsoft
Harrier-OSS-v1, and Jina-v5-text. In every one of them, a linear probe reads
polarity with high accuracy---up to $91.3\%$---while the cosine geometry barely
distinguishes polarity at all: the mean similarity margin between same-polarity
and opposite-polarity sentence pairs is between $0.003$ and $0.064$. The
information is present and linearly accessible; it is simply not metrically
expressed. We call this the \emph{decodability--geometry dissociation}, and we
quantify it with a reusable diagnostic we term the \textbf{Attribute Alignment
Gap} (AAG).

The dissociation matters precisely because production systems consume the metric.
A pipeline that retrieves customer reviews for ``what do users dislike about this
product'' will, with a state-of-the-art embedder, return topically relevant but
polarity-mixed neighbours, because topic dominates the geometry and polarity does
not participate in it. Scaling the encoder does not fix this. Qwen3-Embedding-4B,
one of the strongest open embedders available, has the highest probe accuracy in
our panel but one of its largest probe--MAP gaps ($91.3\%$ probe versus $64.3$
MAP; only Arctic-large is larger), consistent with added capacity improving
semantic geometry without improving affective geometry.

Why does this happen? Contrastive pretraining objectives are trained on pairs
whose relatedness is topical: NLI premises and hypotheses, titles and bodies,
queries and documents. Under such objectives, \emph{``this film is superb''} and
\emph{``this film is dreadful''} are near-optimal positives---identical in
syntax, topic, entities, and register, differing in a single evaluative token.
The objective therefore actively \emph{compresses} the polarity direction rather
than expanding it. We verify that this is structural rather than a prompting
artifact using a control that recent task-conditioned encoders make possible: the
same Jina-v5 weights, conditioned on its \texttt{classification} task instead of
its \texttt{retrieval} task, recover only about half of what injection recovers.

Finally, we show the dissociation is cheaply repairable, and that this reframes a
line of work begun five years ago. The original \emph{Sent2Vec}
model~\cite{sent2vec2020} argued that sentiment should be an explicit component of
a sentence vector, and built one by weighting GloVe vectors with Smoothed Inverse
Frequency (SIF). We retain that thesis but reinterpret the mechanism: a small
supervised-contrastive projection head on a \emph{frozen} modern encoder does not
add sentiment information---the probe results show the information was already
there---it makes latent sentiment metrically accessible, at a measured cost to general
semantic similarity (Section~\ref{sec:cost}). Injection is \textbf{metric
repair} for one attribute, and a re-purposing of the space for that attribute.

Our contributions:
\begin{itemize}
\item We identify and systematically measure a decodability--geometry
dissociation for sentiment across eleven frozen encoders spanning 2021--2026,
including 2026-generation and 4B-parameter LLM embedders, and show it does not
diminish with encoder scale or recency.
\item We contribute the \textbf{Attribute Alignment Gap}, a label-only,
model-agnostic diagnostic that quantifies how metrically expressed any discrete
attribute is in any embedding space, computable in a few lines of code.
\item We isolate the cause with a task-conditioning control on fixed weights,
showing explicit task specialization recovers only about half of the MAP gain
injection achieves and leaves AAG unchanged, which points to the training
objective rather than the inference mode.
\item We show a small supervised head trained in seconds on a CPU repairs the
metric for sentiment, improving every post-2022 encoder on probing and retrieval,
and we measure its cost on general semantics (STS-B).
\item We release a fully CPU-reproducible
pipeline\footnote{Code, raw results, and configuration:
\url{https://github.com/mahdinaser/sent2vec-sise}} covering all eleven encoders.
\end{itemize}

\section{Related Work}
\subsection{Sentence embeddings and their geometry}
Static word embeddings~\cite{mikolov2013word2vec,pennington2014glove} supported
strong sentence representations by weighted averaging; the SIF
scheme~\cite{arora2017sif} remains a hard baseline. SBERT~\cite{reimers2019sbert}
adapted BERT~\cite{devlin2019bert} into a siamese encoder, and
SimCSE~\cite{gao2021simcse} established contrastive learning as the dominant
recipe. Weakly-supervised contrastive pretraining produced
E5~\cite{wang2022e5} and BGE~\cite{xiao2023bge}; Arctic-Embed~\cite{merrick2024arctic}
refined the data recipe; and decoder-based LLMs are now converted into
state-of-the-art encoders, including LLM2Vec~\cite{llm2vec2024} and
Qwen3-Embedding~\cite{qwen3embedding2025}, tracked by
MTEB~\cite{muennighoff2023mteb}. A parallel line studies the \emph{geometry} of
these spaces---anisotropy, alignment and uniformity~\cite{wang2020alignment}---but
characterizes semantic structure. We instead show that a specific, practically
important attribute is systematically excluded from the metric while remaining
linearly decodable, and that this persists in 2026-generation encoders.

\subsection{Sentiment-aware representations}
Sentiment-specific word embeddings~\cite{tang2014sswe} first observed that generic
embeddings place antonymous sentiment terms too close together---a lexical
precursor of the sentence-level effect we quantify. SentiBERT~\cite{yin2020sentibert}
and SentiLARE~\cite{ke2020sentilare} inject sentiment via pretraining or
fine-tuning. Closest to us is SentiCSE~\cite{senticse2024}, which pretrains a
sentiment-aware sentence encoder with a dual word- and sentence-level contrastive
objective and proposes a sentiment-guided textual-similarity benchmark. Our aims
differ in kind: SentiCSE builds a better sentiment encoder, whereas we
\emph{diagnose} a property of general-purpose encoders that practitioners are
already deploying, quantify it with an intrinsic geometric measure that requires
no similarity-annotated benchmark, and use a deliberately minimal frozen-encoder
intervention as evidence that the deficit is metric rather than informational.
Sent2Vec~\cite{sent2vec2020} is the direct ancestor of the intervention;
supervised contrastive learning~\cite{khosla2020supcon} is the objective we adopt.

\section{Diagnostic: the Attribute Alignment Gap}
Let $\mathcal{D}=\{(x_i,y_i)\}_{i=1}^{n}$ be sentences with discrete attribute
labels, $f$ a frozen encoder, and $\bar{\mathbf{e}}_i=f(x_i)/\lVert
f(x_i)\rVert_2$. Define
\begin{equation}
\mathrm{AAG}(f,\mathcal{D}) =
\!\!\mathop{\mathbb{E}}_{\substack{i<j \\ y_i=y_j}}\!\!\left[\bar{\mathbf{e}}_i^{\!\top}\bar{\mathbf{e}}_j\right]
-
\!\!\mathop{\mathbb{E}}_{\substack{i<j \\ y_i\neq y_j}}\!\!\left[\bar{\mathbf{e}}_i^{\!\top}\bar{\mathbf{e}}_j\right].
\end{equation}
AAG is the margin by which the space's own metric separates the attribute. It
needs only categorical labels---no graded similarity annotations---applies to any
attribute and any encoder, and is $O(n^2)$ in similarity evaluations but trivially
cheap on test-set scales. $\mathrm{AAG}\approx 0$ means the metric is blind to the
attribute regardless of how decodable it is.

We pair AAG with standard \emph{linear-probe accuracy}, which measures
decodability. The two together separate two questions that the literature usually
conflates: \emph{is the attribute represented?} (probe) and \emph{is it
metrically expressed?} (AAG). A large probe--AAG discrepancy is the dissociation
this paper reports.

\section{Method: injection as metric repair}
\subsection{Head and objective}
Given a frozen embedding $\bar{\mathbf{e}}\in\mathbb{R}^{d}$, SISE learns
\begin{equation}
g_\theta(\bar{\mathbf{e}}) = \frac{W_2\,\sigma(W_1\bar{\mathbf{e}}+b_1)+b_2}
{\lVert W_2\,\sigma(W_1\bar{\mathbf{e}}+b_1)+b_2\rVert_2},
\end{equation}
with $\sigma$ the ReLU non-linearity, a $256$-unit hidden layer, and output
dimension $k=128 \ll d$. The head has $256d + 33{,}152$ parameters: about
$0.13$M for $d=384$, $0.30$M for $d=1024$ and $0.69$M for $d=2560$. Training uses the supervised contrastive
loss~\cite{khosla2020supcon}: for a batch $B$ with $\mathbf{u}_i=g_\theta(\bar{\mathbf{e}}_i)$
and positives $P(i)=\{p\in B: y_p=y_i, p\neq i\}$,
\begin{equation}
\mathcal{L} = \sum_{i\in B}\frac{-1}{|P(i)|}\sum_{p\in P(i)}
\log\frac{\exp(\mathbf{u}_i^{\!\top}\mathbf{u}_p/\tau)}
{\sum_{a\in B, a\neq i}\exp(\mathbf{u}_i^{\!\top}\mathbf{u}_a/\tau)},
\end{equation}
with $\tau=0.07$. The primary representation is $g_\theta(\bar{\mathbf{e}})$; a
\emph{fused} variant $[\bar{\mathbf{e}};g_\theta(\bar{\mathbf{e}})]$ retains
general semantics.

\subsection{Why this is repair, not addition}
The interpretation matters. Because the base probe accuracy is already high
(Table~\ref{tab:main}), polarity is linearly present in $\bar{\mathbf{e}}$ before
any training; a linear read-out exists. What the space lacks is a metric in which
that direction carries weight. The head is a learned re-metrization: it discards
directions that dominate cosine similarity but are irrelevant to the attribute,
and rescales the low-variance attribute direction until it governs proximity. This
predicts three things we then observe: injection should raise AAG by an order of
magnitude; it should transform few-shot behaviour, since a metrically clean space
needs almost no labels to place a boundary; and it should help \emph{most} in
retrieval, which consumes the metric directly, and least in full-data probing,
which never needed the metric. Note that the head itself is always trained on the
full labelled training set ($4{,}000$ examples); the few-shot experiments below
restrict only the downstream probe.

\subsection{Evaluation protocol}
All representations are frozen at evaluation. We report (i) full-data
linear-probe accuracy, (ii) few-shot probe accuracy at $k\in\{2,5,10,50,100\}$
labels per class averaged over $10$ samplings (for SISE, the probe sees $k$
labels per class but the head was trained on all $4{,}000$, so this measures
few-shot \emph{readout} after supervised training, not label-matched few-shot
learning), (iii) sentiment retrieval by
leave-one-out cosine ranking, reporting precision@$k$ and MAP with respect to
shared polarity, and (iv) AAG. Features are standardized and probes are
$\ell_2$-regularized logistic regressions.

\section{Experimental Setup}
\textbf{Data.} SST-2~\cite{socher2013sst} ($4{,}000$-sentence training subsample,
$872$-sentence validation split used as test, since official test labels are
withheld; head hyperparameters were fixed in advance, and the ablation in
Section~\ref{sec:results} on this split is reported for transparency, not used
for selection) and MR~\cite{pang2005mr} in its Rotten Tomatoes form ($4{,}000$-sentence
training subsample, $1{,}066$ test). An identical protocol is applied to every
encoder so results are directly comparable.

\textbf{Encoders.} Eleven frozen encoders (twelve configurations, counting
Jina-v5-small's \texttt{classification} mode) plus a classical baseline, spanning
2021--2026 and about $22$M--$4$B parameters: MiniLM (all-MiniLM-L6-v2, $384$d);
E5-large-v2, BGE-large-en-v1.5, mxbai-embed-large-v1, Snowflake-Arctic-Embed-L
(all $1024$d); Qwen3-Embedding-0.6B ($1024$d) and
Qwen3-Embedding-4B ($2560$d)~\cite{qwen3embedding2025};
Microsoft Harrier-OSS-v1 at $270$M ($640$d) and $0.6$B
($1024$d)~\cite{harrier2026}; Jina-v5-text-nano ($768$d) and
Jina-v5-text-small ($1024$d)~\cite{jinav5}, the latter additionally under its
\texttt{classification} task; and GloVe+SIF ($100$d) as a Sent2Vec-style
reference. Jina-v5 requires an explicit task, which we exploit as a control
(Section~\ref{sec:control}); unless stated otherwise we use \texttt{retrieval},
the general-purpose deployment setting.

\textbf{Implementation.} Heads train for $60$ epochs with Adam (lr $10^{-3}$,
weight decay $10^{-5}$), batch $512$, $\tau=0.07$, on cached embeddings only.
Everything---including embedding all eleven encoders and the 4B model---runs on a
2-core CPU with no GPU; each head trains in seconds.

\section{Results}
\label{sec:results}

\begin{table*}[t]
\centering
\caption{The decodability--geometry dissociation and its repair, across eleven
frozen encoders (2021--2026; the \emph{cls.\ task} row is Jina-v5-small in
classification mode). \textsc{probe} = full-data linear-probe accuracy;
\textsc{map} = sentiment-retrieval mean average precision; $\Delta$ = Attribute
Alignment Gap. Base columns show the dissociation: high \textsc{probe}, low
\textsc{map}, $\Delta\!\approx\!0$ everywhere. \textbf{Bold} marks injected values
exceeding the base encoder. Only the 2021 MiniLM regresses on \textsc{probe}.}
\label{tab:main}
\setlength{\tabcolsep}{4.4pt}
\footnotesize
\begin{tabular}{llccccccc@{\hskip 10pt}cccccc}
\toprule
& & & \multicolumn{6}{c}{\textbf{SST-2}} & \multicolumn{6}{c}{\textbf{MR}} \\
\cmidrule(lr){4-9}\cmidrule(lr){10-15}
& & & \multicolumn{3}{c}{base} & \multicolumn{3}{c}{$+$ SISE} & \multicolumn{3}{c}{base} & \multicolumn{3}{c}{$+$ SISE} \\
\cmidrule(lr){4-6}\cmidrule(lr){7-9}\cmidrule(lr){10-12}\cmidrule(lr){13-15}
Encoder & Yr & dim & probe & MAP & $\Delta$ & probe & MAP & $\Delta$ & probe & MAP & $\Delta$ & probe & MAP & $\Delta$ \\
\midrule
GloVe+SIF        & --   & 100  & 74.7 & 54.4 & .024 & -- & -- & -- & 73.0 & 52.9 & .016 & -- & -- & -- \\
\midrule
MiniLM           & '21 & 384  & 77.4 & 54.8 & .015 & 76.0 & \textbf{66.2} & \textbf{.131} & 76.3 & 52.8 & .008 & 73.5 & \textbf{62.6} & \textbf{.106} \\
E5-large         & '22 & 1024 & 89.7 & 72.2 & .023 & \textbf{91.9} & \textbf{90.1} & \textbf{.409} & 82.9 & 65.7 & .016 & \textbf{87.2} & \textbf{82.3} & \textbf{.325} \\
BGE-large        & '23 & 1024 & 86.8 & 73.0 & .060 & \textbf{91.3} & \textbf{88.3} & \textbf{.428} & 83.8 & 67.6 & .044 & \textbf{87.1} & \textbf{81.6} & \textbf{.317} \\
mxbai-large      & '24 & 1024 & 87.0 & 73.0 & .064 & \textbf{91.1} & \textbf{88.4} & \textbf{.428} & 84.1 & 67.6 & .047 & \textbf{85.6} & \textbf{80.8} & \textbf{.314} \\
Arctic-large     & '24 & 1024 & 82.2 & 54.4 & .005 & \textbf{86.1} & \textbf{79.6} & \textbf{.264} & 79.2 & 52.6 & .003 & \textbf{80.8} & \textbf{72.7} & \textbf{.195} \\
Qwen3-0.6B       & '25 & 1024 & 84.2 & 61.9 & .027 & \textbf{88.3} & \textbf{83.9} & \textbf{.364} & 81.1 & 58.3 & .019 & \textbf{84.8} & \textbf{77.2} & \textbf{.258} \\
Qwen3-4B         & '25 & 2560 & 91.3 & 64.3 & .029 & \textbf{92.9} & \textbf{90.3} & \textbf{.393} & 85.4 & 60.9 & .023 & \textbf{88.7} & \textbf{84.3} & \textbf{.385} \\
Harrier-270m     & '26 & 640  & 86.6 & 69.4 & .030 & \textbf{89.8} & \textbf{85.3} & \textbf{.387} & 85.9 & 71.6 & .034 & \textbf{88.7} & \textbf{83.5} & \textbf{.355} \\
Harrier-0.6B     & '26 & 1024 & 90.8 & 73.6 & .034 & \textbf{93.3} & \textbf{91.8} & \textbf{.512} & 85.4 & 75.6 & .041 & \textbf{90.8} & \textbf{85.8} & \textbf{.394} \\
Jina-v5-nano     & '26 & 768  & 82.0 & 61.6 & .035 & \textbf{87.8} & \textbf{83.5} & \textbf{.256} & 80.6 & 57.8 & .023 & \textbf{83.4} & \textbf{76.5} & \textbf{.250} \\
Jina-v5-small    & '26 & 1024 & 86.2 & 63.2 & .039 & \textbf{89.9} & \textbf{85.1} & \textbf{.313} & 79.9 & 58.7 & .026 & \textbf{83.8} & \textbf{77.6} & \textbf{.267} \\
\;\;\emph{cls.\ task} & '26 & 1024 & 83.9 & 73.8 & .043 & \textbf{87.5} & \textbf{81.5} & \textbf{.305} & 79.7 & 68.4 & .032 & \textbf{84.1} & \textbf{76.9} & \textbf{.249} \\
\bottomrule
\end{tabular}
\end{table*}

\subsection{The dissociation is universal and scale-invariant}
The base columns of Table~\ref{tab:main} contain the paper's central observation.
Probe accuracy is high and improves markedly across generations---from $77.4\%$
(MiniLM, 2021) to $91.3\%$ (Qwen3-4B, 2025) and $90.8\%$ (Harrier-0.6B, 2026) on
SST-2. AAG, meanwhile, stays pinned near zero for \emph{every} encoder:
$\Delta\in[0.003,0.064]$, with no upward trend over five years of progress and
across a roughly $180\times$ range of parameter counts. Retrieval MAP tracks AAG rather
than probe accuracy, as the metric interpretation predicts.

\begin{figure}[t]
\centering
\includegraphics[width=\linewidth]{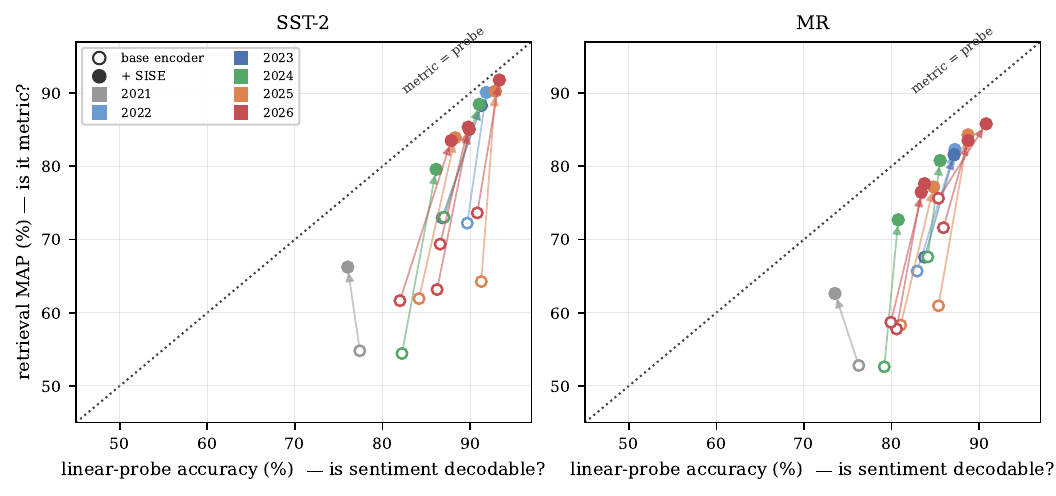}
\caption{Decodability (x) versus metric expression (y). On the dotted diagonal,
the metric would express polarity as well as a probe decodes it. All base
encoders (hollow) fall far below it regardless of year or scale; arrows show where
injection (filled) moves them. Colour encodes model generation.}
\label{fig:quadrant}
\end{figure}

Fig.~\ref{fig:quadrant} makes the dissociation visual. If polarity were as
metrically expressed as it is decodable, encoders would lie on the dotted
diagonal. Every base encoder (hollow markers) sits far below it. Qwen3-4B is a
stark case: the highest base probe accuracy in the panel ($91.3\%$) paired with
a MAP of $64.3$. (Probe accuracy and MAP measure different tasks, so their
difference is an informal indicator rather than a calibrated measure; on it,
Arctic-large at $27.8$ is marginally above Qwen3-4B at $27.0$.) Scaling capacity
does not, by itself, improve affective geometry.

\subsection{Why: the attribute is dilute, not hidden}
\label{sec:variance}
The dissociation has a simple quantitative explanation. Let $\mathbf{w}$ be the
unit class-mean-difference direction and define the \emph{polarity variance
share} $\rho$ as the fraction of an embedding space's total variance lying along
$\mathbf{w}$. On SST-2, $\rho$ is $2.4\%$ for MiniLM, $4.3\%$ for Qwen3-4B,
$6.6\%$ for Harrier-0.6B, and $6.7\%$ for E5-large. Because cosine similarity is
a normalized sum over all $d$ coordinates, an attribute holding four percent of
the variance cannot steer the metric no matter how cleanly a probe reads it: the
signal is \emph{dilute}, not hidden. Scale does not concentrate it---the $4$B
model is more dilute than the $0.6$B one---which is precisely why AAG fails to
improve across generations. Injection acts directly on this quantity, raising
$\rho$ to $85.5\%$--$96.7\%$; the metric changes because the variance budget
changes. Complete variance-accounting measurements, including behaviour under
lossy compression, are released with our code.

\subsection{The deficit is objective-level, not a prompting artifact}
\label{sec:control}
\begin{figure}[t]
\centering
\includegraphics[width=\linewidth]{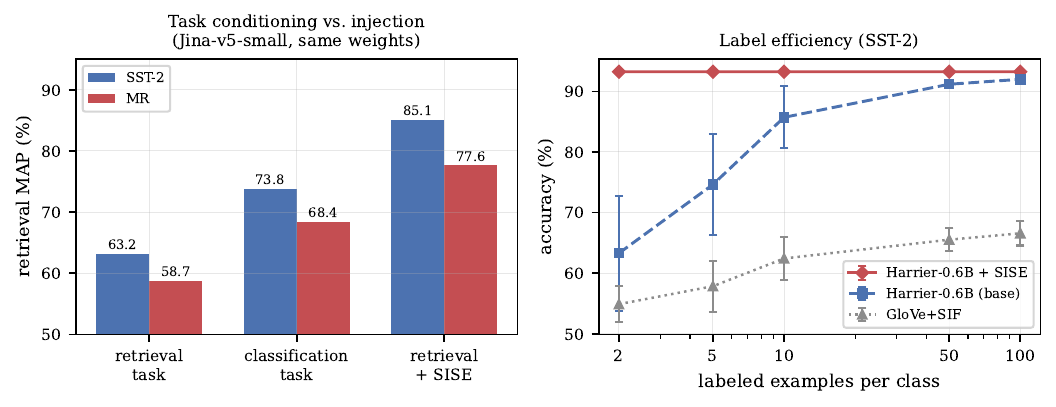}
\caption{Left: task-conditioning control on identical Jina-v5-small weights.
Explicit \texttt{classification} conditioning recovers only about half of the
MAP gain that injection achieves in \texttt{retrieval} mode. Right: probe label
efficiency on a 2026 encoder; after the head is trained on the full labelled set,
the probe needs $\sim$2 labels per class to reach its ceiling.}
\label{fig:control}
\end{figure}

A natural objection is that these encoders are merely being queried in the wrong
mode, and that task conditioning would supply the missing geometry.
Task-conditioned 2026 encoders let us test this directly on \emph{fixed weights}.
Conditioning Jina-v5-small on its \texttt{classification} task rather than
\texttt{retrieval} does improve affective geometry---MAP $63.2\rightarrow73.8$ on
SST-2 and $58.7\rightarrow68.4$ on MR---confirming that task conditioning is not
inert. But injection on the \emph{same weights in retrieval mode} reaches $85.1$
and $77.6$ respectively (Table~\ref{tab:main}, Fig.~\ref{fig:control} left). Task
specialization thus recovers roughly half of the MAP gain that injection achieves
($10.6/21.9 = 48\%$ on SST-2, $9.7/18.9 = 51\%$ on MR), and leaves $\Delta$
essentially unchanged ($0.039\rightarrow0.043$). Notably,
injection still adds $+7.7$ MAP on top of the classification-conditioned variant.
This points to what the pretraining objective optimizes rather than to how the
encoder is addressed at inference, though a task switch on one model cannot fully
isolate pretraining objectives across models.

\subsection{Repair improves every post-2022 encoder, in every regime}
Injection raises AAG by $7$--$61\times$ (largest for Arctic-Embed, whose base
$\Delta=0.003$ is the most affect-blind space we measure) and lifts retrieval MAP
by $+9.8$ to $+26.0$ points. Every encoder from 2022 onward also gains on
full-data probing, by $+1.4$ to $+5.8$ points; only the 2021 MiniLM regresses
($-1.4$, $-2.7$), consistent with its weaker base representation having less
latent structure for a $128$-dimensional head to preserve.

The largest probe effects are in the low-label readout regime. With the head
trained on the full labelled set and the probe restricted to two labels per
class, SISE gains up to $+31.8$ points over the base encoder's 2-shot probe
(Arctic, SST-2), $+29.9$ (Harrier-0.6B), and $+29.6$ (Qwen3-4B), and reaches
essentially its own full-data accuracy, whereas base probes need $50$--$100$
labels per class (Fig.~\ref{fig:control}, right). This comparison is not
label-matched: the SISE pipeline uses $4{,}000$ labels in total. It shows that in
a repaired space the boundary is trivially placed; it does not show that SISE is a
few-shot learner. Training the head itself on $k$ labels per class is left to
future work.

\subsection{Robustness and ablations}
Table~\ref{tab:robust} retrains every head with five seeds: standard deviation is
at most $1.0$ point on SST-2 across all eleven neural encoders and metrics. The
fused variant is consistently at least as accurate as compact SISE on full-data
probing, making it the safer default when general semantics must be retained.
Ablating the head on E5-large/SST-2, results are flat for $k\in\{64,128,256\}$
($92.3/92.4/92.6$ probe) and for $\tau\in\{0.05,0.07,0.1\}$
($92.4/92.4/92.1$), degrading only at an over-smooth $\tau=0.2$ ($91.7$).
No careful tuning is required.

\begin{table}[t]
\centering
\caption{Five-seed robustness of injection on SST-2 (mean$_{\pm\text{std}}$, \%).
Fused = SISE-fused full-data probe accuracy. MR behaves identically
(std $\leq1.3$).}
\label{tab:robust}
\footnotesize
\setlength{\tabcolsep}{4pt}
\begin{tabular}{lcccc}
\toprule
Encoder & probe & fused & 2-shot & MAP \\
\midrule
MiniLM        & $76.7_{\pm0.5}$ & $76.9_{\pm0.6}$ & $76.1_{\pm0.8}$ & $66.4_{\pm0.3}$ \\
E5-large      & $92.4_{\pm0.3}$ & $92.5_{\pm0.1}$ & $92.2_{\pm0.4}$ & $90.2_{\pm0.3}$ \\
BGE-large     & $91.1_{\pm0.4}$ & $91.4_{\pm0.4}$ & $90.9_{\pm0.2}$ & $88.1_{\pm0.3}$ \\
mxbai-large   & $91.4_{\pm0.4}$ & $91.6_{\pm0.5}$ & $91.2_{\pm0.4}$ & $88.5_{\pm0.2}$ \\
Arctic-large  & $85.2_{\pm0.9}$ & $85.5_{\pm0.9}$ & $85.1_{\pm0.9}$ & $79.5_{\pm0.9}$ \\
Qwen3-0.6B    & $88.6_{\pm0.6}$ & $88.7_{\pm0.6}$ & $88.4_{\pm0.4}$ & $83.8_{\pm0.3}$ \\
Qwen3-4B      & $92.3_{\pm0.3}$ & $92.8_{\pm0.5}$ & $92.4_{\pm0.2}$ & $90.1_{\pm0.5}$ \\
Harrier-270m  & $89.7_{\pm0.6}$ & $89.9_{\pm0.4}$ & $89.6_{\pm0.2}$ & $85.4_{\pm0.3}$ \\
Harrier-0.6B  & $\mathbf{93.3}_{\pm0.2}$ & $\mathbf{93.4}_{\pm0.1}$ & $\mathbf{93.2}_{\pm0.2}$ & $\mathbf{91.5}_{\pm0.1}$ \\
Jina-v5-nano  & $87.9_{\pm0.7}$ & $88.7_{\pm0.3}$ & $88.2_{\pm0.3}$ & $83.7_{\pm0.2}$ \\
Jina-v5-small & $89.3_{\pm0.3}$ & $89.9_{\pm0.2}$ & $89.6_{\pm0.3}$ & $85.8_{\pm0.3}$ \\
\bottomrule
\end{tabular}
\end{table}

\subsection{The dissociation is not specific to sentiment}
\label{sec:attrs}
If the deficit really originates in the pretraining objective rather than in
something peculiar to affect, it should appear for other attributes. We test this
directly on nine labelled attributes spanning five families---topic (AG News),
affect (tweet sentiment, tweet emotion, six-way emotion), pragmatics
(subjectivity, irony), social (offensiveness, hate speech), and stance (climate)
---six of which share a single text domain (\textsc{TweetEval}), so domain is
held constant. Table~\ref{tab:attrs} reports, for three encoders, the probe lift
over the majority-class baseline alongside AAG.

The pattern is uniform. Probes recover these attributes easily---up to $+62.9$
points of lift for topic and $+42.8$ for subjectivity---while AAG never exceeds
$0.08$ and is usually below $0.05$, against $0.4$--$0.5$ in a repaired space
(Table~\ref{tab:main}). Strikingly, this includes \emph{topic}: the one property
contrastive objectives are explicitly trained to encode still registers an AAG of
only $0.02$--$0.08$. Sentiment is therefore not a special case; it sits at the
low end of an effect that spans every attribute family we measured.

Two controls are worth noting. First, attributes that probes \emph{cannot} decode
---offensiveness and hate speech, with lifts of $0.3$ and $-1.6$ on
Harrier-0.6B---also show no metric structure. Those are representational
absences, not dissociations, and any AAG study must check decodability before
claiming one. Second, we do \emph{not} observe a monotone relationship between
encoder quality and AAG across this panel: E5-large records the lowest gaps and
MiniLM the highest, with the 2026 Harrier-0.6B in between. The robust claim is
universality across attributes, not an ordering across encoders.

\begin{table}[t]
\centering
\caption{The dissociation across nine attributes and three encoders. \emph{lift}
= linear-probe accuracy minus majority-class baseline (Harrier-0.6B, 2026);
AAG columns give the alignment gap for each encoder. Probes read these attributes
well; the metric expresses none of them.}
\label{tab:attrs}
\footnotesize
\setlength{\tabcolsep}{4pt}
\begin{tabular}{llcc@{\hskip 8pt}ccc}
\toprule
 & & & lift & \multicolumn{3}{c}{AAG} \\
\cmidrule(lr){5-7}
Attribute & Family & $K$ & (Har.) & MiniLM & E5-L & Har. \\
\midrule
Topic (AG News)  & topic     & 4 & $+62.9$ & .080 & .020 & .043 \\
Subjectivity     & pragmatic & 2 & $+42.8$ & .045 & .015 & .050 \\
Emotion (6-way)  & affect    & 6 & $+35.6$ & .017 & .009 & .015 \\
Emotion (tweet)  & affect    & 4 & $+33.4$ & .034 & .013 & .023 \\
Stance (climate) & stance    & 3 & $+10.7$ & .054 & .014 & .026 \\
Sentiment (tweet)& affect    & 3 & $+9.7$  & .016 & .005 & .008 \\
Irony            & pragmatic & 2 & $+4.6$  & .011 & .003 & .002 \\
Offensiveness    & social    & 2 & $+0.3$  & $-$.004 & .001 & $-$.004 \\
Hate speech      & social    & 2 & $-1.6$  & .006 & .002 & .003 \\
\bottomrule
\end{tabular}
\end{table}

\subsection{What the repair costs}
\label{sec:cost}
A $128$-d head trained only on polarity should be expected to discard other
structure, and it does. On STS-B (Spearman), the compact SISE output scores
$44.4$, $47.0$ and $42.4$ for MiniLM, E5-large and Harrier-0.6B, against $82.0$,
$87.7$ and $82.5$ for the base embeddings: a loss of $38$--$41$ points. The fused
variant $[\bar{\mathbf{e}};g_\theta(\bar{\mathbf{e}})]$ retains most of it
($79.6$, $75.7$, $78.1$; a loss of $2.4$--$12.0$). Compact SISE is therefore a
sentiment-specialized index, not a general replacement; the fused variant is the
appropriate choice when one index must serve both.

\subsection{Big-Data consequence}
Injection is a $128$-dimensional output regardless of $d$, trained in seconds on a
CPU over embeddings computed once. Two deployment implications follow. First,
sentiment specialization can be added to an existing embedding index without
re-encoding the corpus or fine-tuning the encoder. Second, once both are given
the same head, encoder scale matters little for sentiment retrieval:
\textbf{Harrier-0.6B $+$ SISE reaches $91.8$ MAP and Qwen3-4B $+$ SISE reaches
$90.3$}, both at $128$ dimensions, with $6.7\times$ fewer encoder parameters for
the former. Compared with the unsupervised $2560$-d Qwen3-4B embedding ($64.3$
MAP), a $128$-d sentiment index is $20\times$ smaller -- at billion-vector scale,
roughly $10$\,TB of float32 versus about $512$\,GB -- but that comparison mixes
supervised and unsupervised representations and should be read as a footprint
argument, not as a like-for-like quality comparison.

\subsection{Discussion}
Read together, the results argue that embedding quality is not one quantity.
Five years of progress delivered large gains in semantic geometry and in
decodability, and essentially none in affective geometry; AAG is as flat for a
2026 4B-parameter LLM embedder as for a 2021 MiniLM.
A caveat applies to AAG itself: it is an absolute cosine margin, and its scale
depends on each encoder's anisotropy, so equal AAG values are not strictly
comparable across spaces, and a small AAG does not by itself rule out good
nearest-neighbour ranking. AAG should be read alongside retrieval MAP, as we do
throughout; a normalized variant is a natural refinement. This is a blind spot with a
plausible cause---antonymous sentence pairs are near-perfect positives under
topical contrastive objectives---an inexpensive diagnostic (AAG), and an
inexpensive repair. It also suggests a general practice: before deploying an
embedding index for any attribute-sensitive task, measure AAG for that attribute
rather than assuming leaderboard quality implies metric coverage.

\subsection{Limitations}
Our study covers two English binary-sentiment benchmarks with $4{,}000$-example
training subsamples, eleven encoders, and a linear-probe protocol; we do not
fine-tune encoders, and the largest models we could run on a CPU-only budget are
$4$B parameters, leaving $8$--$12$B leaders untested. The repair experiments
target one attribute; Section~\ref{sec:attrs} shows the underlying dissociation
holds for eight further attributes, but we do not train repair heads for them, so
whether injection transfers as cleanly beyond polarity remains open. The
attribute panel also uses three encoders rather than eleven, and two of its
attributes (offensiveness, hate speech) prove undecodable in these encoders and
so contribute controls rather than evidence. We explicitly do \emph{not} claim that the benefit of
injection grows with encoder quality: across eleven encoders, probe-accuracy gains
are approximately flat ($+1.4$ to $+5.8$) above MiniLM and \emph{smaller} for the
two strongest models, which sit nearer the ceiling. What is robust is
universality, not monotone scaling. We also do not compare against simpler
supervised projections (e.g.\ LDA or rescaling along the probe direction), against
SentiCSE, or against instruction-prompted encoders given a sentiment instruction;
these baselines would show how much of the gain requires a non-linear contrastive
head. The few-shot results use a head trained on all labels (Section~\ref{sec:results}). Finally, encoder ``quality'' and output
dimension are partly confounded in any public model panel; a dimension-controlled
study (e.g.\ Matryoshka truncations of one model) would isolate the two.

\section{Conclusion}
We showed that modern text embedders---through 2026-generation and 4B-parameter
LLM-based models---encode sentence sentiment in a form that linear probes read
almost perfectly and their own metric almost ignores, with the largest
dissociation appearing in the strongest model tested. We introduced the Attribute
Alignment Gap to quantify this, demonstrated with a fixed-weight task-conditioning
control that the deficit is not merely a prompting artifact, and
showed that a small supervised-contrastive head repairs the metric for sentiment
on a CPU in seconds---improving every post-2022 encoder's probe accuracy and
retrieval MAP (up to $+26.0$), at a measured cost to general semantic similarity
that a fused representation largely avoids. The
thesis of the original Sent2Vec---that sentiment belongs explicitly in the
vector---survives the LLM transition, but its justification changes: the goal is
not to add information the encoder lacks, but to repair a metric that hides
information the encoder already has. Future work includes testing the
dissociation for other attributes, compositional multi-attribute repair in
orthogonal subspaces, cross-lingual transfer, and weakly supervised repair with
LLM-generated labels for corpora without annotations.

\section*{Reproducibility}
Embedding extraction for all eleven encoders, the contrastive head, and the
probe / few-shot / retrieval / AAG evaluation use only open datasets and open
models and run end-to-end on a commodity 2-core CPU. Seeds are fixed throughout.
Code and raw results: \url{https://github.com/mahdinaser/sent2vec-sise}.

\end{document}